# Evidence-Bound Reasoning: Neuro-Semantic Verification of Biomedical AI in Glioblastoma Radiogenomics

Mariya Miteva[1], Maria Nisheva-Pavlova[2]

*Faculty of Mathematics and Informatics – Sofia University St. Kliment Ohridski, 5 James Bourchier Blvd., Sofia 1164, Bulgaria*

## Abstract

Background: Biomedical artificial intelligence (AI) can generate plausible explanations without providing a reliable mechanism to verify whether each statement is supported by patient-specific evidence. We developed a neuro-semantic verification framework that converts quantitative radiomic measurements into addressable evidence records and machine-checkable claims. Methods: University of Pennsylvania Glioblastoma (UPenn-GBM) radiomics were aligned with de novo Cancer Imaging Phenomics Toolkit (CaPTk) extraction from standardized magnetic resonance imaging (MRI) and expert-validated segmentations in an independent multicenter cohort. The shared space comprised 1,728 features from T1-weighted (T1), contrast-enhanced T1-weighted (T1GD), T2-weighted (T2), and fluid-attenuated inversion recovery (FLAIR) MRI across three tumor regions. We derived reference-defined semantic states from 611 UPenn cases and evaluated cross-cohort transportability, model-linked provenance, deterministic claim verification, controlled predictive degradation, and a large language model (LLM) claim-extraction pilot; O6-methylguanine-DNA methyltransferase (MGMT) prediction served only as a transport stress test. Results: Median semantic-state agreement was 0.786 (weighted kappa 0.709), ranging from 0.918 for morphologic to 0.252 for intensity features. Our external evidence ledger contained 1,655 model-linked records for 331 patients. Our verifier achieved 100% exact-set accuracy in a 6,620-claim compound-corruption benchmark. In a 24-case pilot, GPT-5.6 Sol reproduced 72/72 prespecified atomic claims, after which the frozen verifier recovered 24/24 expected verification conditions. During controlled degradation, receiver operating characteristic area under the curve (ROC AUC) declined from 0.899 to 0.500 while verification accuracy remained 1.000. External MGMT discrimination was weak (ROC AUC 0.543). Conclusions: Our results show that verifiability can be engineered and evaluated independently of predictive performance. LLMs may structure narrative explanations, while final evidence-consistency checking remains deterministic.

Keywords: glioblastoma; radiomics; radiogenomics; verifiable artificial intelligence; explainable artificial intelligence, evidence provenance; model auditing; large language models

## Introduction

AI in medical imaging is increasingly expected to provide explanations in addition to predictions. Yet explanation and verification are not equivalent. A post-hoc narrative can be coherent, persuasive, and medically plausible while still failing to identify the exact patient-specific evidence that supports each statement. This distinction is especially important when models are transported between institutions, where scanner characteristics, acquisition protocols, preprocessing, segmentation, and cohort composition can change the numerical distributions on which the model was developed [4-8,13]. In that setting, a readable explanation may describe the model output without demonstrating that the explanation is faithful to the evidence actually available for the patient.

Within explainable artificial intelligence (XAI), many approaches emphasize interpretability, feature attribution, or post-hoc rationales, but these outputs do not by themselves establish that each explanatory statement is faithful to the specific evidence available for an individual patient. In healthcare, this limitation is particularly consequential: Ghassemi et al. [13] highlighted fundamental limitations of current explainability approaches, while the broader hallucination literature has shown that fluent natural-language outputs can remain unsupported by source evidence [14]. More recent healthcare-specific work further underscores the need for evidence-grounded safeguards in clinical AI systems [31]. However, these strands do not by themselves provide a patient-level, machine-checkable contract that resolves each explanatory claim to a uniquely addressable quantitative evidence record. The present work is designed to address that specific gap.

[1] Corresponding author. Tel.: +359-899-661-730
*E-mail address*: mmiteva@fmi.uni-sofia.bg
[2] Tel.: +359-879-243-434
*E-mail address*: marian@fmi.uni-sofia.bg

Radiogenomics provides a useful stress test for this problem. Glioblastoma (GBM) is spatially heterogeneous, and tissue-based molecular assessment samples only part of the tumor. MRI captures complementary phenotypic information through enhancement, morphology, signal distribution, edema or infiltration, diffusion, and perfusion. Radiomics converts these patterns into quantitative descriptors, whereas radiogenomics investigates associations between imaging phenotype and molecular characteristics [4-8]. MGMT promoter methylation is clinically relevant because of its relationship to temozolomide response and outcome [3], but imaging-based MGMT prediction has shown substantial methodological and cross-cohort variability [7,8]. That variability makes MGMT useful here not as the central claim, but as a deliberately difficult transport setting in which to test whether an evidence pathway remains auditable when prediction is uncertain.

Conventional radiomic pipelines represent a patient as a high-dimensional numerical feature vector. Such vectors preserve quantitative detail but do not naturally encode the semantic units used in biomedical reasoning. In prior work, we introduced a controlled neuro-semantic representation in which radiomic measurements were translated into patient-level phenotype tokens retaining modality, tumor region, semantic domain, and cohort-relative state [12]. That study established the representation concept and showed that hierarchical and hybrid semantic representations could support molecular modeling and evidence-grounded explanations. Here, we extend that representation from an interpretable language into a formal evidence architecture designed for machine-checkable biomedical reasoning.

Compared with that earlier neuro-semantic representation [12], the present study adds three elements that change the role of the representation: addressable Evidence IDs with explicit provenance, a formal atomic-claim schema, and a deterministic verification layer. The goal is therefore not simply to make a model more interpretable, but to make the evidentiary relation between a statement and its patient-specific source explicitly testable. This distinction also defines the boundary with causal AI. The proposed framework does not estimate causal effects or infer biological mechanisms from radiomic associations; instead, it treats unsupported causal language as a formally detectable violation of evidentiary scope.

We introduce evidence-bound reasoning as a formal design in which quantitative measurements are converted into addressable neuro-semantic evidence objects with explicit provenance and a fixed claim schema. In our approach, a formal evidence record links a patient-specific measurement to its imaging sequence, tumor region, exact CaPTk descriptor, semantic domain, relative state, reference thresholds, and, when applicable, model contribution. A downstream explanation may remain human-readable prose, but before verification we convert it into machine-readable atomic claims that cite Evidence IDs and normalize modality, region, semantic state, model probability, and interpretive scope. An LLM may perform this text-to-schema conversion, but it does not determine whether the claim is true. Our deterministic verifier resolves the structured claim against the stored evidence record, avoiding a circular design in which one language model generates biomedical prose and another unconstrained language model is asked to judge its correctness.

Our approach explicitly separates four properties that are often conflated: predictive validity, semantic transportability, language-to-schema fidelity, and claim verifiability. We used two public glioblastoma cohorts to construct an exact shared radiomic space, quantified how reference-defined semantic states transported across institutions, built model-linked patient evidence ledgers, added a proof-of-concept LLM interface for converting narrative statements into atomic claims, and stress-tested our rule-based verifier with both compound claim corruption and controlled predictive degradation. The central question was not whether a fluent explanation could be produced, but whether every clinically meaningful statement could be reduced to an inspectable claim and checked against patient-specific evidence.

## Proposed Evidence-Bound Reasoning Framework

Figure 1 summarizes the complete architecture as a six-stage evidence pathway. First, the reference and external cohorts define the empirical setting. Second, both cohorts are aligned in an exact shared structural-radiomic space so that evidence identity is fixed before interpretation. Third, quantitative measurements are converted into neuro-semantic evidence states while retaining the reference thresholds from which those states were derived. Fourth, predictive model output is kept as a separate stress-test layer rather than being treated as evidence of explanation validity. Fifth, readable narrative statements can be normalized by an LLM into atomic claim objects. Sixth, a frozen deterministic verifier resolves each claim against the patient-specific evidence ledger and checks identity, imaging context, semantic state, model output, and interpretive scope.

This organization is intentionally different from a conventional XAI pipeline in which an attribution or generated rationale is itself presented as the explanation. Here, explanation is an interface layer, whereas verifiability is an independently testable system property. The same separation is important with respect to causal AI: model-linked feature contributions are retained as descriptors of fitted model behavior, but they are not promoted to causal effects. Claims that cross that boundary are rejected by the verifier. The three outputs shown at the bottom of Figure 1 - transportable evidence, machine-readable claims, and deterministic verification - therefore correspond to distinct evaluation targets rather than a single aggregate notion of 'explainability'.

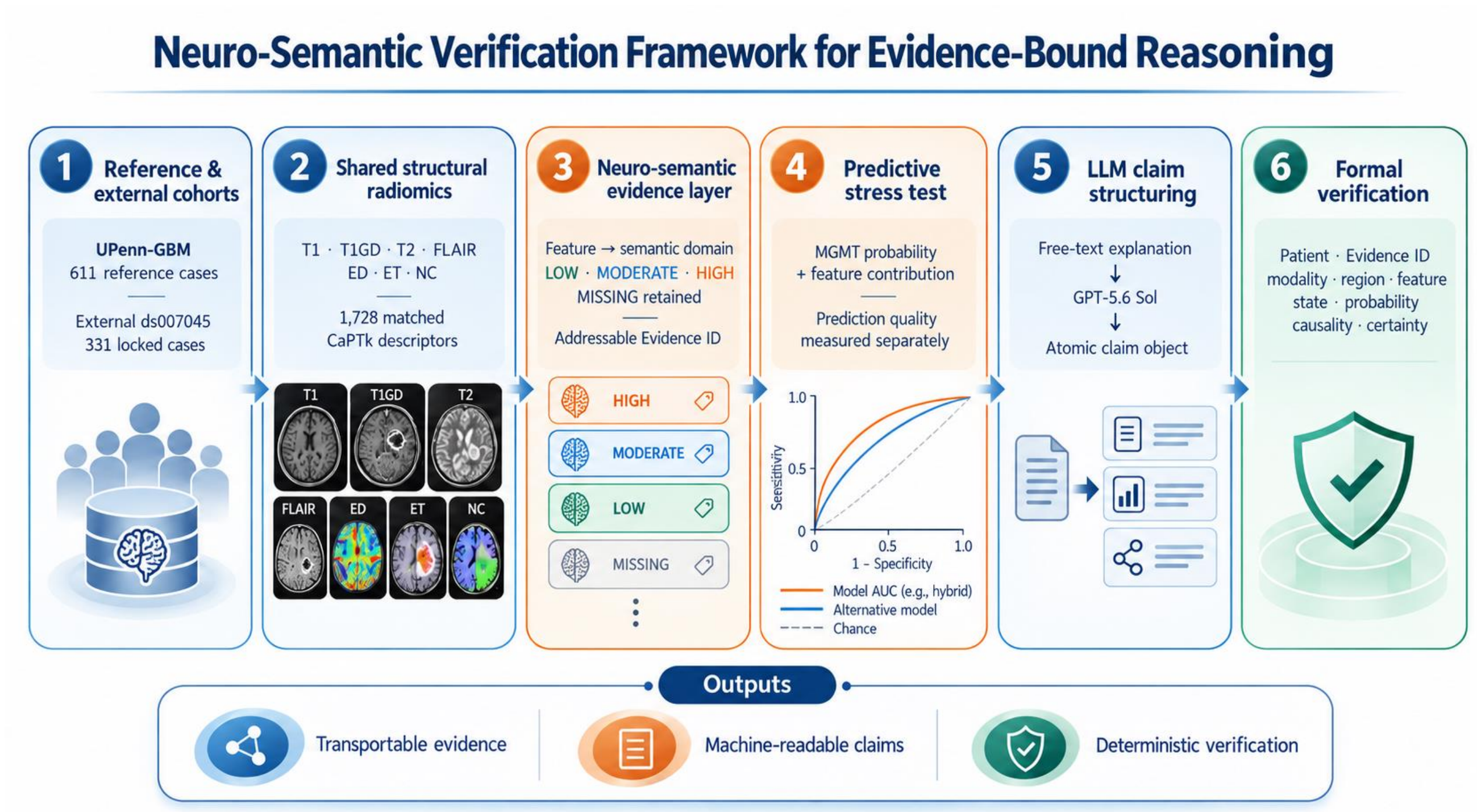

Figure 1. Neuro-semantic verification framework for evidence-bound biomedical AI reasoning. Two glioblastoma cohorts are aligned in an exact shared structural-radiomic space; quantitative measurements are converted into reference-defined semantic evidence records with addressable Evidence IDs. MGMT prediction is retained as a separate stress-test output. Human-readable explanations may be structured by an LLM into atomic claim objects, while final consistency checking remains deterministic. The lower strip summarizes the three system outputs evaluated in this study: transportable evidence, machine-readable claims, and deterministic verification.

# Methods

## Study design and prespecified analysis hierarchy

This retrospective computational study used public de-identified glioblastoma datasets and was organized around a prespecified hierarchy of endpoints. The primary methodological endpoint was claim-level verifiability: whether a biomedical statement encoded in the formal claim schema could be resolved to patient-specific evidence and checked for exact consistency. The secondary endpoint was transportability of the neuro-semantic evidence layer across cohorts. MGMT discrimination was a tertiary empirical stress test used to expose the evidence system to realistic predictive failure rather than to establish a new diagnostic biomarker. A final proof-of-concept interface experiment evaluated whether a contemporary general-purpose LLM could convert schema-constrained biomedical prose into the atomic claim object required by the verifier. This LLM pilot was explicitly exploratory and was not used to define verifier rules or predictive models.

The final analysis was version-locked before manuscript reconstruction. Cohort definitions, the 1,728-feature dictionary, reference thresholds, semantic mappings, predictive model configuration, top-evidence selection rule, and verifier rules were held fixed. External outcomes were not used to redesign the ontology or choose semantic thresholds. The controlled degradation experiment was added as a systems stress test after the formal verification rules had been frozen; it was explicitly simulation-based and was not used for biological inference.

Reporting followed the TRIPOD+AI statement by separating model development, external transfer, calibration, representation analysis, and error analysis [17]. The study was additionally informed by the Checklist for Artificial Intelligence in Medical Imaging (CLAIM), the DECIDE-AI reporting guideline, and the FUTURE-AI consensus guideline, particularly their emphasis on transparent evaluation pathways, traceability, robustness, and explainability [20-22]. These frameworks were used as reporting and design references rather than as claims of prospective clinical validation. The source resources were public and de-identified; no new patient contact or intervention occurred. Ethics and consent procedures are those reported by the source datasets [9,16].

## Reference and independent cohorts

The UPenn-GBM cohort served as the reference dataset [9,10]. Baseline preoperative radiomics generated with the CaPTk were available for 611 cases and distributed through The Cancer Imaging Archive (TCIA). Clinical v2.1 data were linked by exact identifier; 262 of the 611 radiomic cases had definitive binary MGMT status, including 111 methylated and 151 unmethylated tumors. Earlier neuro-semantic work used a previous clinical-data version and a 278-case MGMT subset [12]. The present analysis therefore constitutes a new version-locked representation and verification study rather than a numerical replication of the earlier predictive results.

The independent cohort was the multicenter glioblastoma MRI dataset reported by Filimonova et al. [16]. The published cohort comprised 337 patients from eight hospitals with standardized preoperative structural MRI, expert-reviewed tumor segmentation, and MGMT annotation. Because ds007045 did not provide a radiomic matrix directly compatible with the UPenn-GBM CaPTk schema, we extracted the external radiomic features de novo from the standardized imaging derivatives and segmentation masks. Extraction was completed for all 363 available ds007045 records. Exact matching of the resulting radiomic table against the published v1.0.1 participant set yielded 331 patients with aligned radiomics and MGMT status: 150 methylated and 181 unmethylated. We excluded the six published identifiers without matching radiomic rows and did not replace them with newer radiomic-only patients.

## De novo radiomic feature extraction for the independent cohort

Radiomic features for the independent ds007045 cohort were generated specifically for this study using CaPTk v1.9.0 FeatureExtraction. The standardized preprocessed MRI derivatives distributed with ds007045 were used directly. These comprised skull-stripped, N4 bias-field-corrected, z-score-normalized volumes registered to the Montreal Neurological Institute 152 (MNI152) template: T1, T1GD, T2, and FLAIR MRI. The corresponding expert-validated Brain Tumor Segmentation (BraTS)-format masks defined enhancing tumor, necrotic/non-enhancing tumor core, and peritumoral edema. No clinical, molecular, survival, or outcome variable entered the radiomic extraction workflow.

Features were extracted independently for every MRI modality-tumor-region combination. The extraction schema and feature ordering were fixed a priori to reproduce the UPenn-GBM CaPTk structural-radiomic definitions used by the locked development pipeline. Exactly 144 CaPTk descriptors were generated for each modality-region pair, corresponding to 432 descriptors per MRI modality and 1,728 descriptors per subject across four modalities and three tumor compartments. Feature names and ordering were harmonized to the UPenn-GBM CaPTk feature list before any model application or semantic encoding. This emphasis on fixed feature identity and explicit provenance is consistent with broader radiomics standardization efforts such as the Image Biomarker Standardisation Initiative, while the present study did not assume that software-level feature-name correspondence alone guaranteed cross-site numerical equivalence [23].

The extraction pipeline treated anatomical absence and numerical failure explicitly. If a tumor subregion was absent, its features were stored as missing rather than as zero-valued measurements, thereby avoiding interpretation of non-existent anatomy as a quantitative radiomic signal. Infinite values produced during extraction were retained in the raw extraction output for traceability and converted to missing values only in the model-ready table. When minor image-mask geometry discrepancies were encountered, segmentation masks were resampled into the corresponding image space using nearest-neighbor interpolation so that discrete BraTS label values were preserved.

This de novo extraction step was deliberately outcome-blind and served only to reconstruct the shared quantitative evidence space required for technical transportability testing. Compatibility was assessed at the level of exact feature identity: downstream analyses used only descriptors whose names and positions matched the locked UPenn structural-radiomic inventory, with no outcome-guided remapping or retrospective feature substitution. Because multicenter MRI radiomics remains sensitive to scanner, acquisition, preprocessing, and segmentation effects, numerical compatibility was evaluated separately from transportability rather than treated as evidence of harmonization [24]. This separation also follows contemporary radiomics-quality recommendations that emphasize transparent preprocessing, harmonization assessment, external validation, and failure-mode analysis [25]. The cohort definitions, radiomics provenance, molecular-label availability, and exact shared representation used in our analysis are summarized in Table 1.

Table 1. Cohorts and shared evidence representation. T1GD, contrast-enhanced T1-weighted MRI; ED, edema/infiltrative region; ET, enhancing tumor; NC, necrotic core.

| Characteristic | UPenn-GBM reference/development | Independent multicenter cohort |
|---|---|---|
| **Radiomic cases** | 611 baseline cases | 331 aligned published cases |
| **Radiomics provenance** | Precomputed CaPTk structural radiomics supplied with UPenn-GBM | Extracted de novo in this study with CaPTk v1.9.0 from ds007045 derivatives |
| **Definitive MGMT** | 262 (111 methylated / 151 unmethylated) | 331 (150 methylated / 181 unmethylated) |
| **Structural MRI** | T1, T1GD, T2, FLAIR | T1, T1GD, T2, FLAIR |
| **Tumor regions** | ED, ET, NC | ED, ET, NC |
| **Exact shared radiomic features** | 1,728 | 1,728 |
| **Role in study** | Reference semantics; model development | Transport, evidence, and audit stress test |

### Exact radiomic feature alignment and missingness

Both datasets were reduced to the exact intersection of structural-MRI CaPTk feature names. The final shared space contained 1,728 descriptors: four MRI sequences x three tumor regions x 144 descriptors per modality-region pair. No approximate feature-name matching, feature remapping, or outcome-guided feature reconciliation was used. The shared inventory included 732 histogram, 240 intensity, 228 morphologic, 24 volumetric, 96 gray-level co-occurrence matrix (GLCM), 120 gray-level run-length matrix (GLRLM), 216 gray-level size-zone matrix (GLSZM), 60 neighboring gray-tone difference matrix (NGTDM), and 12 local binary pattern (LBP) descriptors [11].

Feature completeness was high but not perfect. Of the 611 UPenn cases, 597 were complete across all 1,728 descriptors; 301 of 331 external cases were complete. Patients were not excluded solely for isolated missing measurements. The semantic evidence layer retained missingness as an explicit MISSING state instead of silently assigning LOW, MODERATE, or HIGH. This was also preserved in the model-linked evidence ledger so that absence of a quantitative measurement remained inspectable.

### Neuro-semantic evidence layer

For every shared feature, the 25th percentile (Q25), median, and 75th percentile (Q75) reference values were estimated from all 611 UPenn radiomic cases without using MGMT labels. External values below Q25 were encoded LOW, values above Q75 HIGH, and values between Q25 and Q75 MODERATE; absent measurements were encoded MISSING. These states are reference-relative descriptors and were not interpreted as universal biological cut-offs.

Feature families were mapped to controlled semantic domains while preserving imaging context. Intensity and histogram descriptors were represented as intensity-distribution evidence; morphologic and volumetric descriptors represented tumor morphology. Texture features were interpreted according to modality: T1 texture as T1 signal heterogeneity, T1GD texture as enhancement heterogeneity, T2 texture as T2 signal heterogeneity, and FLAIR texture as infiltration-associated heterogeneity. Each patient-feature instance was assigned a unique Evidence ID and stored with patient identifier, modality, tumor region, exact descriptor, feature family, semantic domain, state, raw value when present, and UPenn Q25/Q75 provenance.

We introduce the Evidence ID as the formal mechanism that transforms an interpretable neuro-semantic token into an auditable patient-specific evidence object. A phrase such as T1GD_ET_HIGH_ENHANCEMENT_HETEROGENEITY is readable but cannot by itself prove which measurement generated it. In our framework, an addressable record resolves the semantic phrase to a unique patient, feature, reference distribution, and model-linked contribution. The resulting evidence layer functions as a provenance contract connecting quantitative radiomics to any downstream explanatory claim.

### Cross-cohort semantic transportability

Transportability was evaluated independently of MGMT. Every external measurement was first encoded with the locked UPenn Q25/Q75 thresholds. For diagnostic comparison only, Q25/Q75 values were also calculated within the external cohort, and the locally recalibrated state was compared with the UPenn-locked state for the same patient-feature measurement. Agreement was summarized feature-wise and by radiomic family. Linearly weighted Cohen kappa was used to preserve the ordinal relation among LOW, MODERATE, and HIGH states.

Distribution shift was additionally quantified in the raw and semantic spaces. For each descriptor, the two-sample Kolmogorov-Smirnov statistic summarized quantitative shift, while total-variation distance summarized change in the semantic-state distribution. These metrics were interpreted as representation diagnostics rather than evidence of biological invariance. A separate sensitivity analysis used the manually corrected UPenn segmentation subset to determine whether the overall transport pattern was driven primarily by the automatic segmentation pipeline.

### Transport diagnostics and segmentation sensitivity

Transportability was examined at three complementary resolutions. At the measurement level, every external patient-feature pair could be assigned a locked reference state and, when the quantitative measurement was present, a locally recalibrated state. At the feature level, agreement summarized the proportion of external patients whose semantic category was unchanged by reference transfer. At the family level, the median of these feature-specific agreements was used to characterize broad representation behavior without allowing large feature families to dominate simply because they contained more descriptors. The same hierarchy was applied to weighted kappa.

The manual-segmentation sensitivity analysis used 232 UPenn cases with manually corrected radiomics, including 228 complete cases across the shared feature space. Reference quantiles were recomputed from this subset and compared with the same 331 external patients. The purpose was not to search for a more favorable reference distribution, but to test whether the observed family-specific shift could be explained primarily by the automatic UPenn segmentation. Similar median raw and semantic distribution-shift values under the manual reference would argue that scanner, acquisition, preprocessing, and population effects remained important contributors.

No transportability threshold was treated as a biological truth criterion. The earlier exploratory high-transportability subsets were retained only as sensitivity analyses and were not used to define the evidence ontology. In the final manuscript, transportability is attached to evidence as metadata: a feature may be correctly represented and correctly cited even when its semantic state is known to be unstable across cohorts. This distinction allows the verifier to answer whether a claim is faithful without pretending that all evidence is equally transferable.

### Empirical MGMT transport stress test

MGMT prediction was retained as a secondary empirical stress test because it exposes the evidence architecture to a clinically meaningful but difficult cross-cohort task. Fixed Light Gradient Boosting Machine (LightGBM) classifiers were evaluated for raw, semantic, and hybrid structural-radiomic inputs [15]. The model configuration was deliberately constrained and held constant: 100 trees, learning rate 0.05, seven leaves, maximum depth 3, minimum child size 15, feature subsampling 0.4, L2 regularization 10, L1 regularization 2, balanced class weights, and random seed 42. Development robustness was assessed with three independently shuffled stratified five-fold evaluations; no hyperparameter search against the external cohort was performed.

For locked transfer, semantic thresholds were estimated once from all 611 UPenn reference cases without labels. The classifier was fitted only on the 262 cases with definitive MGMT status and applied unchanged to the 331 external patients. External prediction was summarized by receiver operating characteristic area under the curve (ROC AUC), average precision, Brier score, and calibration slope. This analysis was not designated the primary endpoint of the manuscript; its purpose was to provide a real domain-shift setting in which the audit system could be evaluated under predictive uncertainty.

### Model-linked evidence attribution

For every external patient, feature-contribution values from the transferred semantic model were calculated. The five largest absolute contributions were retained as the compact patient-specific evidence set. Each record stored the signed contribution and contribution direction together with the evidence provenance fields. Contribution scores were interpreted as model-behavior descriptors, not causal effects.

We designed the final external evidence ledger as a compact auditing representation containing five ranked evidence records for each of 331 patients. This fixed-size design prioritizes verifiability over explanation maximalism: an explanation can remain concise while every cited imaging statement remains addressable. Our ledger also stores feature-level transportability, allowing a reviewer to distinguish a correctly cited but domain-sensitive measurement from a more stable one.

### LLM-mediated conversion of free-text explanations to atomic claims

The formal verifier begins with a structured claim object, whereas clinical-facing explanations are naturally expressed as prose. To test the technical handoff between these layers, a fixed 24-case interface pilot was constructed from external-cohort evidence records. Four cases were sampled from each of six prespecified narrative conditions: fully supported evidence, semantic-level mismatch, region mismatch, modality mismatch, unsupported causal interpretation, and molecular overstatement. Each narrative contained two patient-specific imaging statements derived from model-linked evidence records and one statement reporting the model-estimated MGMT probability. Lexical variants were used for imaging sequences and tumor compartments (for example, 'post-contrast T1-weighted imaging' for T1GD and 'peritumoral edema/infiltrative compartment' for ED), so the extraction step required normalization rather than simple field copying. The resulting benchmark contained 24 narratives and 72 atomic claims.

GPT-5.6 Sol (OpenAI; ChatGPT, High reasoning setting; accessed 13 September 2026) was used in a single-pass claim-structuring pilot [19]. The target schema contained claim type, participant identifier, Evidence ID, modality, tumor region, semantic domain, semantic level, model probability, unsupported-causality flag, and molecular-confirmation flag. The LLM was used only to structure the statement; it was not used as the final verifier. Structured outputs were scored against the known benchmark claim objects using exact field match, exact atomic-claim match, and exact case-level match, and the extracted claims were then passed unchanged to the frozen deterministic verifier. BioMistral-7B, Meditron-70B, and Mistral-7B-Instruct-v0.3 were not rerun in the present experiment; they are relevant comparator models for a future multi-model claim-structuring benchmark and had been used as biomedical/general LLM comparators in the preceding neuro-semantic explanation study [12,26-28]. The broader medical-LLM literature further supports separating clinical-language capability from factual verification [29]. Schema-constrained generation is also conceptually aligned with grammar-constrained decoding approaches that enforce machine-readable output structures, although no claim is made that the interactive ChatGPT interface used here implements the specific decoding algorithm of those systems [30]. Because the narratives were study-authored, schema-constrained, and evaluated with a single model in an interactive environment in which decoding parameters are not exposed, this experiment was designated a technical interface-feasibility pilot rather than a blinded or comparative LLM benchmark.

## Evidence-object contract and verification rule engine

The evidence ledger was designed as an explicit contract rather than a display-only explanation table. Required identity fields included patient, Evidence ID, modality, tumor region, feature ID, and exact feature name. Interpretation fields included semantic domain and semantic state. Provenance fields included the UPenn reference Q25/Q75 and reference-cohort identifier. Model-linkage fields included probability, signed contribution, and contribution direction. Raw value was stored when available; if the measurement was absent, the MISSING state itself became the auditable evidence condition.

We designed the verifier as a set of local, composable rules that independently test evidence identity, imaging context, feature identity, semantic state, model output, and interpretive scope. Identity rules check that the cited record exists and belongs to the expected patient; context rules compare modality and tumor region; feature rules check the exact CaPTk feature identifier and descriptor; semantic rules check the controlled domain and state; and model rules check the cited probability and contribution direction. Interpretive-scope rules reject claims that convert association into causality or a model-estimated probability into confirmed molecular status. Because each rule operates on an explicit field, our verifier can return multiple violations simultaneously rather than forcing a single error label.

The verifier was not permitted to infer missing facts from biomedical plausibility. For example, a claim about enhancement heterogeneity could not be accepted merely because an enhancing tumor region existed; the cited Evidence ID had to resolve to the correct modality, region, feature, and state. Similarly, molecular overstatement was rejected even if the model probability was high. This design intentionally prioritizes evidence fidelity over narrative coherence.

## Formal claim schema and compound-corruption benchmark

The verifier operated on atomic structured claims. A claim could specify Evidence ID, patient, modality, region, feature identity, semantic domain, semantic state, model probability, and contribution direction, together with flags for causal interpretation or conversion of a probabilistic estimate into confirmed molecular status. The verifier resolved the Evidence ID and compared each supplied field with the patient evidence ledger.

We constructed a compound-corruption benchmark spanning nine prespecified violation types: semantic-level mismatch, region mismatch, modality mismatch, phenotype invention, feature mismatch, probability mismatch, contribution-sign mismatch, unsupported causal interpretation, and molecular overstatement. The benchmark contained 6,620 claims: 1,655 valid claims and 1,655 claims at each of one, two, or three simultaneous injected violations. Exact-set scoring required our verifier to identify every injected violation and no additional one. This benchmark evaluates the deterministic checking system after claims have already been represented in structured form; it is not a free-text language-model benchmark.

## Controlled predictive-degradation experiment

We designed a controlled predictive-degradation experiment to isolate, experimentally, the relationship between predictive discrimination and formal verifiability. Using the 331 external outcomes solely for simulation, we defined a latent score as S = Y + sigma Z, where Y was the binary MGMT label, Z was a fixed standard-normal noise vector generated with seed 586, and sigma controlled discrimination. Five prespecified noise scales produced realized ROC AUC values of approximately 0.90, 0.80, 0.70, 0.60, and 0.50. Scores were transformed monotonically to probabilities. Because labels were used to construct these scores, this experiment carries no predictive or biological inference; it is an orthogonality test of our verification architecture.

For each discrimination regime and each patient, the top-ranked addressable evidence item from the fixed external evidence ledger was paired with the simulated model probability. Four structured claims were then generated: a supported claim and claims containing one, two, or three simultaneous deterministic violations selected from semantic-level, region, modality, phenotype, feature, contribution-sign, probability, causal, and molecular-overstatement errors. This produced 331 x 5 x 4 = 6,620 claims. The same frozen verifier was applied at every AUC level. The prespecified test was whether exact-set verification accuracy changed as predictive discrimination was progressively degraded.

## Leakage control and reproducibility safeguards

Label-independent semantic reference thresholds were estimated from the radiomic reference population rather than from MGMT-defined subgroups. In development cross-validation, any model-dependent preprocessing was confined to the training fold. The independent cohort was not used to select predictive hyperparameters. The formal verifier rules were defined from the evidence schema, not from observed error cases, and were frozen before both the controlled degradation experiment and the LLM interface pilot. The 24-case language-model pilot did not contribute to ontology design, threshold selection, feature attribution, predictive fitting, or verifier-rule definition.

The controlled degradation simulation was separated analytically from the empirical MGMT model. It used labels only to create predefined discrimination regimes and therefore was never compared as a candidate clinical predictor. The simulation, empirical prediction analysis, evidence ledger, and verifier benchmark were stored as separate result objects so that a high verification score could not be mistaken for evidence of molecular-prediction validity.

### Statistical analysis

Patient-level ROC AUC and average precision summarized predictive discrimination. External AUC confidence intervals (CIs) were estimated with patient-level bootstrap resampling; permutation testing assessed the null hypothesis of chance discrimination in the empirical MGMT stress test. Brier score and calibration slope were retained as descriptive probability metrics. Transportability analyses were outcome-free. State agreement and weighted kappa summarized semantic transport, while raw Kolmogorov-Smirnov and semantic total-variation distances summarized distribution shift.

Verification performance was evaluated by exact-set accuracy. A claim was scored correct only when the complete set of detected violations exactly matched the injected set. The controlled predictive-degradation analysis reports realized AUC, threshold accuracy, balanced accuracy, and exact verification accuracy at each noise regime. For the exploratory LLM interface pilot, field-level exact accuracy, atomic-claim exact match, case-level exact match, and downstream verifier agreement with the prespecified narrative condition were reported descriptively without confidence intervals or inferential testing. No multiplicity-adjusted clinical claims were made because the predictive analyses were secondary stress tests rather than confirmatory biomarker validation.

## Results

### De novo external radiomics reproduced the locked UPenn structural feature schema

The external radiomic matrix was not a source-dataset deliverable but an experiment-specific derivative generated from the standardized ds007045 MRI and expert-validated segmentation masks. CaPTk extraction completed for all 363 available records and produced the prespecified four-modality, three-region structure. After harmonization to the UPenn CaPTk feature list, the expected 144 descriptors were available for each modality-region combination, giving a nominal 1,728-feature structural-radiomic schema per subject.

Version locking was performed only after extraction. The 363-row radiomic matrix was matched to the published v1.0.1 participant set of 337 cases, yielding 331 exact subject matches for the external evidence analysis. The six published participants without matching radiomic rows were excluded, while 32 newer radiomic-only records were retained outside the locked analysis set. This separation ensured that the external cohort remained tied to the published participant definition rather than to the larger extraction inventory.

Quality control preserved missingness rather than forcing numerical completeness. A total of 301 of 331 locked external patients were complete across all 1,728 descriptors; isolated absent measurements and non-finite extraction results were carried forward as missing values and subsequently represented by the explicit MISSING semantic state. Thus the external evidence layer retained both successful measurements and extraction/anatomical absence as auditable states.

### Cohort locking produced an exact 1,728-feature shared evidence space

After de novo extraction and version locking, the structural radiomic intersection was exact: all 1,728 feature names required by the four-sequence, three-region representation were present in the locked inventories of both cohorts. The UPenn reference contained 611 radiomic cases, of which 262 had definitive MGMT labels. The external evidence analysis contained 331 of the 337 published patients after exact identifier alignment. The 32 newer radiomic-only cases were not substituted for the six published patients lacking matching radiomic rows.

This version-locking step was important because the evidence architecture depends on stable feature identity. Each evidence object retained the original CaPTk descriptor rather than a cohort-specific surrogate. Consequently, modality, region, feature family, descriptor identity, threshold provenance, and semantic state were all defined in the same coordinate system across the two datasets.

### Semantic transportability was high for morphology but strongly family dependent

Across the 1,728 features, the median external-patient agreement between UPenn-locked semantic states and locally recalibrated external states was 0.786; the median weighted kappa was 0.709. These aggregate values concealed substantial family-specific heterogeneity. Morphologic features were the most stable, with median state agreement 0.918 and median weighted kappa 0.789. Histogram and GLRLM features also showed relatively high median agreement (0.852 and 0.813, respectively). In contrast, GLCM agreement was 0.328 and first-order intensity agreement was 0.252, with median weighted kappa approximately zero for intensity descriptors.

The distribution-shift analysis showed the same pattern. Median raw Kolmogorov-Smirnov shift was 0.103 for morphology but 1.000 for intensity features. Semantic discretization reduced the intensity-family total-variation distance to 0.750, but this remained a large shift. The representation therefore provided a common syntax and explicit reference frame without making every phenotype state institution invariant. This distinction is important: transportability is a measurable attribute of an evidence item rather than an assumption built into the semantic label. The family-level transportability results are summarized numerically in Table 2 and visualized in Figure 2.

Table 2. Cross-cohort transportability of the neuro-semantic evidence layer by radiomic family.

| Feature family | Features | Median state agreement | Median weighted kappa | Median raw KS | Median semantic TV |
|---|---|---|---|---|---|
| **Morphologic** | 228 | 0.918 | 0.789 | 0.103 | 0.064 |
| **Histogram** | 732 | 0.852 | 0.788 | 0.134 | 0.123 |
| **GLRLM** | 120 | 0.813 | 0.769 | 0.151 | 0.146 |
| **NGTDM** | 60 | 0.792 | 0.742 | 0.179 | 0.165 |
| **Volumetric** | 24 | 0.770 | 0.715 | 0.191 | 0.171 |
| **GLSZM** | 216 | 0.726 | 0.650 | 0.221 | 0.192 |
| **LBP** | 12 | 0.611 | 0.537 | 0.309 | 0.284 |
| **GLCM** | 96 | 0.328 | 0.235 | 0.474 | 0.451 |
| **Intensity** | 240 | 0.252 | 0.000 | 1.000 | 0.750 |

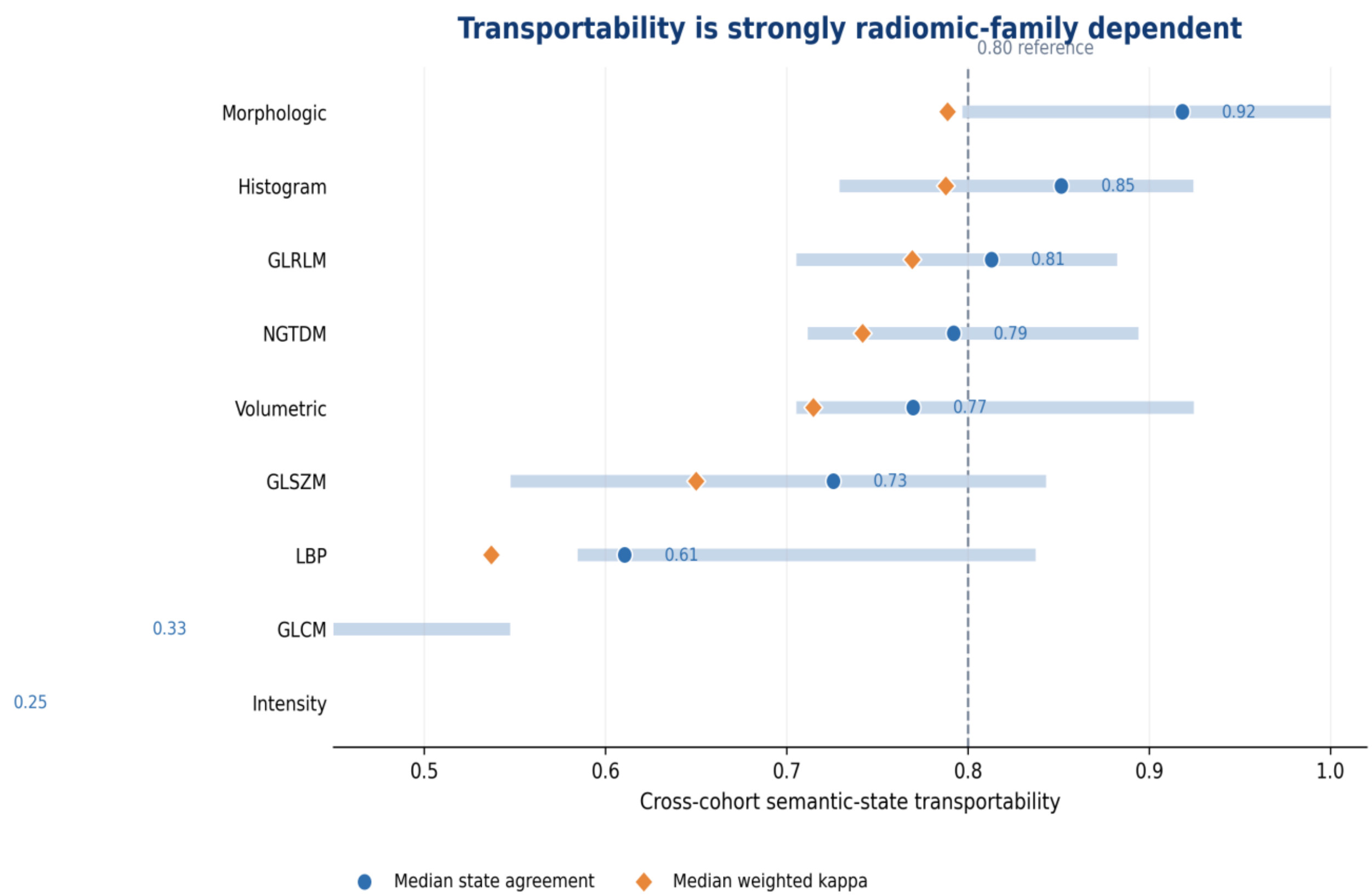

Figure 2. Cross-cohort transportability of neuro-semantic states by radiomic family. Blue circles show median external-patient state agreement and pale horizontal segments show the interquartile range of agreement across features. Orange diamonds show median weighted kappa. The dashed line marks 0.80 agreement. Morphologic, histogram, and GLRLM descriptors were comparatively stable, whereas intensity and GLCM states were strongly reference dependent.

## Distribution-shift sensitivity supported the family-specific transport pattern

Across feature families, semantic compression generally reduced the apparent magnitude of cross-cohort distribution shift but did not remove it. The overall median raw Kolmogorov-Smirnov statistic was approximately 0.181, compared with a median semantic total-variation distance of approximately 0.162. The reduction was most visible for strongly shifted intensity features, where the raw median KS reached 1.000 and semantic total variation was 0.750. By contrast, morphology was relatively stable in both spaces (median raw KS 0.103; semantic total variation 0.064).

The manually corrected UPenn reference produced a similar global pattern: median raw KS 0.176 and median semantic total-variation distance 0.153. The proximity of these values to the automatic-segmentation analysis indicates that the observed domain shift cannot be attributed solely to segmentation choice. The result supports a cautious interpretation of semantic states as reference-anchored evidence categories whose transportability should be measured and reported rather than assumed.

### The evidence ledger made both observed values and missingness explicit

The final model-linked external evidence ledger contained 1,655 records, exactly five for each of 331 patients. Patient identifier, Evidence ID, modality, region, feature identity, semantic domain, reference Q25/Q75, semantic state, model contribution, and contribution direction were available for every record. Forty-five of 1,655 top-evidence records (2.7%) lacked a raw quantitative value and were explicitly represented with the MISSING semantic state. Missingness was therefore visible to the verifier rather than hidden by a narrative explanation.

At the conventional 0.5 threshold of the empirical semantic MGMT model, 181 external predictions were correct and 150 were incorrect. Both groups retained the same evidence schema and five-record provenance structure. Mean semantic-state transportability among the top evidence items was 0.632 in correct cases and 0.650 in incorrect cases (Mann-Whitney p=0.095). Thus an incorrect prediction was not less auditable simply because the label was wrong.

### Model-linked evidence showed a distinct family profile

The transferred semantic model did not use all radiomic families equally among its strongest case-level contributions. Histogram descriptors represented 45.3% of the 1,655 top-five evidence records, intensity descriptors 21.6%, GLSZM 21.3%, GLRLM 10.9%, and morphologic features 0.8%. This profile is a description of model behavior, not a ranking of biological importance. It is nevertheless relevant to audit because the most frequently cited evidence included both relatively transportable histogram features and highly cohort-sensitive intensity features.

Fourteen external patients had at least one top-five record whose quantitative value was absent and therefore carried an explicit MISSING state. This did not prevent provenance tracking: the Evidence ID still resolved to the expected patient, modality, region, descriptor, reference thresholds, and model contribution. In a verifiable system, missingness is itself part of the evidence context rather than a detail that disappears when a narrative explanation is generated.

A representative misclassified methylated case received a model-estimated methylation probability of 0.292. Its strongest cited evidence included low intensity distribution in the T1 edema/infiltrative region, low enhancement heterogeneity in the T1GD necrotic core, and low T1GD intensity distribution in the necrotic core. The molecular prediction was wrong, but the explanatory contract remained enforceable: the system could state exactly which evidence was used while explicitly avoiding a claim that those imaging findings established MGMT status or causality.

### LLM claim structuring provided a machine-readable handoff to the verifier

Our 24-case interface pilot comprised 72 atomic claims: two evidence-linked imaging claims and one model-output claim per narrative. GPT-5.6 Sol reproduced all prespecified schema fields without discrepancy in this constrained benchmark. Exact field accuracy was 1.000 for claim type, patient identifier, Evidence ID, modality, region, semantic domain, semantic level, probability, causality flag, and molecular-confirmation flag. Thus, our interface achieved exact claim-level match for all 72 atomic claims and exact case-level schema reconstruction for all 24 narratives.

The extracted claims were passed directly to the frozen formal verifier without manual correction. The verifier recovered the expected case condition in 24/24 narratives, including all four examples in each of the six categories: supported, semantic-level mismatch, region mismatch, modality mismatch, unsupported causality, and molecular overstatement. This 100% pilot result should be interpreted narrowly. The narratives were study-authored and structurally regular, and the experiment used a single model rather than an independent multi-model benchmark. Its value is to demonstrate that the architecture can place an LLM upstream as a semantic parser while keeping final claim adjudication outside the language model. The resulting language-to-schema handoff and the pilot outcomes are summarized in Figure 3.

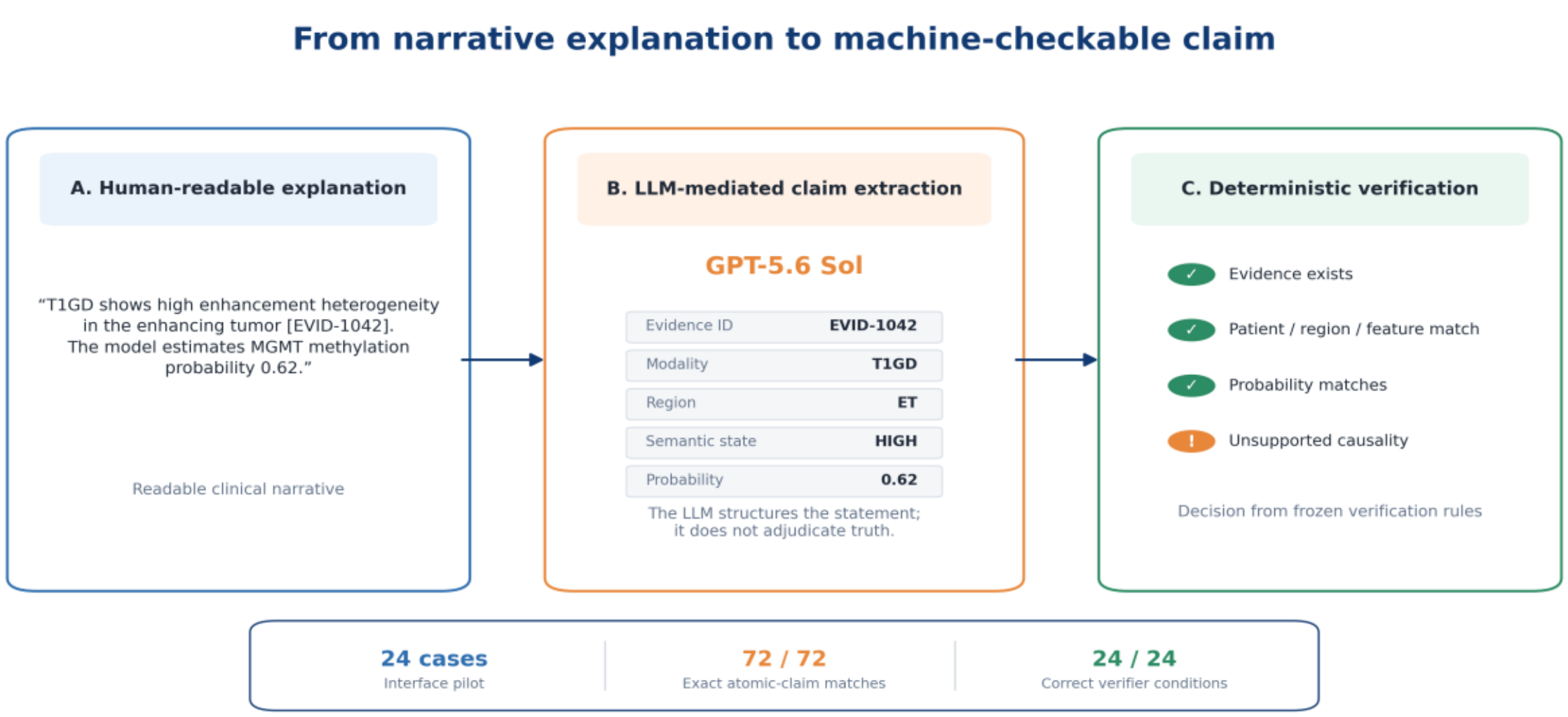

Figure 3. LLM-mediated claim structuring as an interface between readable explanation and formal verification. A human-readable narrative is converted by GPT-5.6 Sol into the prespecified atomic claim schema; the language model structures the statement but does not adjudicate its truth. The frozen deterministic verifier then resolves the structured fields against the patient-specific evidence ledger. In the schema-constrained 24-case pilot, all 72 atomic claims exactly matched the study-authored gold schema and the expected verification condition was recovered in 24/24 cases. These results represent technical interface feasibility rather than independent free-text LLM validation.

## Formal verification remained exact under compound claim corruption

Our verifier recovered the exact injected error set in all 6,620 controlled claims. Exact-set accuracy was 100% for the 1,655 supported controls and remained 100% for claims containing one, two, or three simultaneous violations. The verifier correctly localized errors across semantic level, tumor region, imaging modality, phenotype domain, feature identity, cited probability, contribution direction, unsupported causality, and molecular overstatement.

This result establishes software-level correctness of the formal checking rules under the controlled schema. The separate LLM pilot addresses only the preceding text-to-schema handoff and does not change the interpretation of the 6,620-claim benchmark. Once a statement is represented as an evidence-linked atomic claim, its internal consistency with the patient evidence ledger becomes exactly testable; any language-model error in structuring the claim remains a distinct upstream failure mode rather than being hidden inside the verifier.

Verification performance was identical when claims were stratified by whether the underlying empirical MGMT prediction was correct. The compound benchmark contributed 3,620 claims from correctly classified patients and 3,000 claims from misclassified patients; exact-set detection was 100% in both strata. This is the direct patient-level form of the study's core distinction: the verifier can be correct about what evidence a system used even when the predictor is wrong about the biological target. Claim counts and exact-set performance across increasing corruption depth are summarized in Table 3.

Table 3. Compound claim-verification benchmark. Violation classes included semantic-level, region, modality, phenotype, feature, probability, contribution-sign, unsupported-causality, and molecular-overstatement errors.

| Simultaneous violations | Claims | Exact-set detection |
|---|---|---|
| **0 (supported controls)** | 1,655 | 100.0% |
| **1** | 1,655 | 100.0% |
| **2** | 1,655 | 100.0% |
| **3** | 1,655 | 100.0% |

## Verification was invariant to controlled loss of predictive discrimination

Our controlled predictive-degradation experiment provided the direct test of the study hypothesis. We held cohort size, patient evidence references, claim schema, and frozen verification rules constant while progressively reducing simulated predictive discrimination. Realized ROC AUC values were 0.899, 0.799, 0.700, 0.600, and 0.500. Classification accuracy decreased in parallel from 0.831 to 0.498, confirming that the perturbation altered predictive quality rather than merely calibration.

Exact-set verification accuracy remained 1.000 at every discrimination level and for all zero-, one-, two-, and three-error claims. At the chance-like regime (AUC 0.500), the verifier therefore remained as exact as in the high-discrimination regime (AUC 0.899). Because the simulated scores were deliberately generated from the true labels plus controlled noise, this experiment is not evidence of a new predictor. Its role is to demonstrate system-level orthogonality: the verifier tests whether a claim matches its declared evidence and model output, not whether the model output is clinically correct.

Balanced accuracy fell from 0.833 in the high-discrimination regime to 0.497 in the chance-like regime, whereas formal verification remained exactly 1.000. The decline was monotonic across the five prespecified regimes. This behavior provides a clearer experimental separation than a single weak external AUC: one axis of system quality was deliberately varied over a wide range while the other was measured with the same evidence records and fixed rules.

The controlled claims also preserved compound-error complexity at each discrimination level. Every regime contained 331 supported claims and 331 claims at each of one, two, and three simultaneous violations. Thus stable verification was not driven by easier claims at lower AUC. The same nine error families and exact-set criterion were maintained across the complete discrimination ladder. The full predictive-discrimination ladder is reported in Table 4 and visualized in Figure 4.

Table 4. Controlled predictive-degradation experiment.

| Predictive regime | ROC AUC | Accuracy | Balanced accuracy | Exact verification |
|---|---|---|---|---|
| **High** | 0.899 | 0.831 | 0.833 | 1.000 |
| **Moderate-high** | 0.799 | 0.746 | 0.745 | 1.000 |
| **Moderate** | 0.700 | 0.653 | 0.653 | 1.000 |
| **Weak** | 0.600 | 0.571 | 0.571 | 1.000 |
| **Chance-like** | 0.500 | 0.498 | 0.497 | 1.000 |

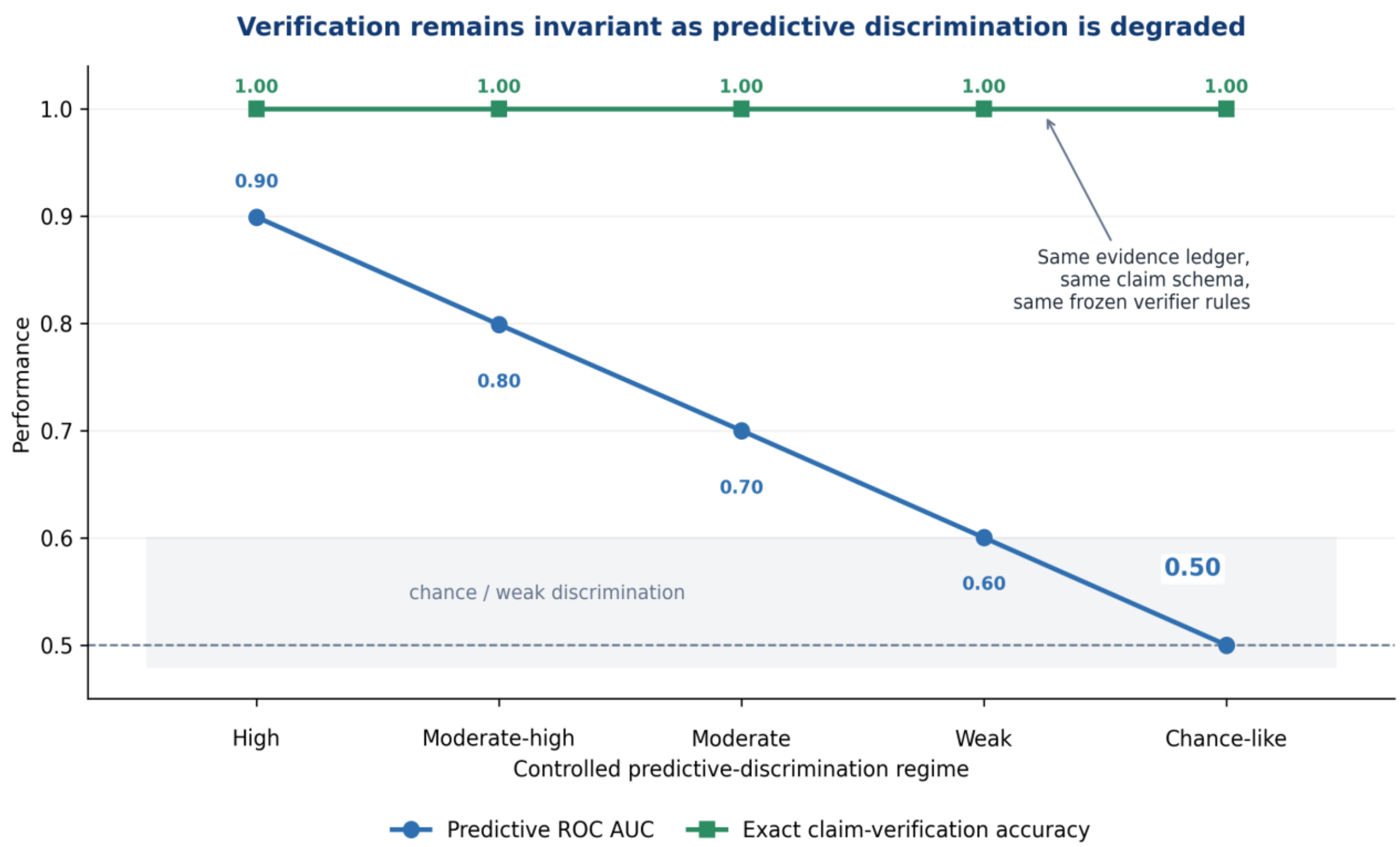

Figure 4. Verification under controlled predictive degradation. Predictive ROC AUC was deliberately reduced from approximately 0.90 to chance while the patient evidence ledger, atomic claim schema, and frozen verification rules were held fixed. Exact-set claim-verification accuracy remained 1.00 at every discrimination level. The experiment isolates verifiability as a system property and does not represent clinical model validation.

### Empirical MGMT transfer supplied a real predictive-failure setting rather than the main endpoint

The locked all-feature semantic MGMT model achieved external ROC AUC 0.543 (95% bootstrap CI 0.478-0.602; permutation p=0.173), average precision 0.487, and Brier score 0.249. The corresponding raw model achieved AUC 0.485. Morphology/volumetry semantics reached AUC 0.538, and the outcome-free 448-feature high-transportability semantic subset reached AUC 0.541. No evaluated external confidence interval excluded 0.5. These analyses therefore do not support external validation of an MGMT predictor.

The negative transport result was retained because it creates a realistic failure mode in which the verification system can be evaluated without conflating a high AUC with trustworthy reasoning. For every real external prediction, including false-positive and false-negative cases, the system still exposed the model-linked evidence records and could reject claims that changed region, modality, state, feature identity, probability, contribution sign, causal meaning, or molecular certainty. The empirical and simulation experiments therefore converged on the same architectural conclusion from different directions: verifiability is not a surrogate for predictive accuracy, and predictive accuracy is not a prerequisite for evidence traceability.

## Discussion

Our central contribution is to show that verifiability can be explicitly engineered and evaluated as a system property distinct from predictive performance. Our formal neuro-semantic layer converts radiomic measurements into addressable evidence objects, preserves reference provenance during cross-cohort transfer, and allows machine-readable claims to be checked exactly against patient-specific evidence. The strongest result of our approach is therefore not an MGMT AUC, but the experimental demonstration that claim verification remained exact under both compound corruption and controlled degradation of predictive discrimination from approximately 0.90 to chance.

This separation resolves an important ambiguity in explainable biomedical AI. A fluent explanation is not itself evidence of fidelity, and clinically capable language models can still produce unsupported or incorrect statements [13,14,29,31]. Likewise, a confident or accurate model output does not demonstrate that a generated explanation refers to the right patient, sequence, tumor region, feature state, or probability. Our architecture makes these relationships explicit by requiring every verifiable statement to resolve to an addressable patient-specific Evidence ID, which in turn resolves to a stored record with defined provenance. Our LLM interface provides the practical bridge from prose to this formal layer: the language model normalizes what the sentence says, while our deterministic verifier decides whether that structured statement agrees with the stored evidence. Unsupported causal interpretation and molecular overstatement therefore become explicit rule violations rather than matters of rhetorical plausibility.

The controlled degradation experiment was deliberately artificial and should be interpreted accordingly. It used the true outcome plus fixed noise to create a broad spectrum of discrimination while holding the evidence and verification machinery constant. Its purpose was not to generate a clinically useful score but to isolate the verification property experimentally. The result formalizes the concept that a verifier can remain correct about the evidentiary relation even when the predictor becomes wrong. The empirical multicenter MGMT analysis then supplied the complementary real-world condition: prediction transferred weakly, yet correct and incorrect cases remained equally addressable and auditable.

The transportability analysis adds a second important qualification. Neuro-semantic encoding did not create a universally invariant vocabulary. Morphologic states were relatively stable across cohorts, while first-order intensity and GLCM states were highly sensitive to the reference distribution. LOW, MODERATE, and HIGH must therefore be interpreted as reference-relative evidence states, not universal biological categories. Their advantage is that the reference threshold is stored with the evidence object. A reviewer can inspect not only what semantic label was assigned but also how that label was defined and how transportable that feature family was in the external cohort.

Our present framework extends the earlier neuro-semantic representation in a critical way: we transform interpretable tokens into addressable evidence objects by adding explicit provenance, model linkage, and deterministic verification [12]. Under evidence-bound reasoning, provenance constrains what a claim is allowed to say: patient, modality, region, descriptor, semantic state, and model output define the permissible evidentiary scope. The formal verifier does not ask whether a sentence sounds medically reasonable; it asks whether its patient, modality, region, descriptor, state, probability, and interpretive scope agree with the evidence ledger. Our design also avoids using a second unconstrained language model as the final judge of the first model's explanation. A future language model can generate human-readable prose and a parallel structured claim object, while the final consistency check remains deterministic [14,18,20-22,30].

The empirical MGMT results remain clinically important precisely because they are weak. The present structural-radiomic model should not be described as an externally validated MGMT predictor. The current clinical v2.1 intersection also differs from the 278-case cohort used in the prior manuscript, and the model configuration was intentionally fixed rather than optimized to reproduce the earlier internal results [12]. Repeated tuning on the current external cohort would compromise the transport experiment. The appropriate next step is a frozen third-cohort evaluation, not further optimization against already inspected outcomes.

As a second methodological contribution, we generated the independent radiomic matrix de novo from standardized ds007045 MRI derivatives rather than relying on a pre-existing external feature table. Unlike UPenn-GBM, where CaPTk radiomics were distributed with the dataset, the external cohort required reconstruction of the quantitative evidence layer. Our extraction pipeline reproduced the locked CaPTk feature definitions using fixed modality-region decomposition, explicit handling of absent regions and non-finite values, and exact feature-name harmonization. This reduced a major source of ambiguity in cross-cohort radiomics and is consistent with literature emphasizing standardized feature definitions, transparent preprocessing, harmonization assessment, and reproducible external validation [23-25]. At the same time, our de novo extraction does not eliminate domain shift: scanner, acquisition, preprocessing, segmentation, and cohort differences remain encoded in the values. For this reason, we report semantic-state transportability and threshold provenance rather than assuming that numerical compatibility implies biological equivalence.

Positioned in the broader XAI landscape, evidence-bound reasoning complements rather than replaces attribution, visualization, or natural-language explanation. Its specific contribution is to add an executable evidence contract beneath those interfaces: every verifiable statement must resolve to a patient-specific record with explicit provenance. This also clarifies the relationship to causal AI. The framework does not infer interventions, counterfactual effects, or biological causal mechanisms; instead, it enforces a boundary between associative/model-linked evidence and causal language by making unsupported causal interpretation a detectable claim violation. Accordingly, explainability, causality, predictive validity, and evidentiary verifiability should be evaluated as related but non-interchangeable system properties.

Several limitations should be emphasized. The study is retrospective and uses two cohorts for the empirical transport analysis. The external evidence experiment included 331 of 337 published patients because six identifiers lacked matching radiomic rows. The shared feature space was restricted to structural MRI and therefore did not test the diffusion and perfusion information available in UPenn-GBM. Feature contributions describe fitted model behavior rather than biological causality. The LLM claim-structuring experiment was small (24 cases), study-authored, schema-constrained, non-comparative, and not a blinded independent benchmark; its 100% exact-match result should therefore be interpreted as interface feasibility rather than evidence of general free-text extraction performance. Only GPT-5.6 Sol was evaluated for claim structuring in the present study. BioMistral-7B, Meditron-70B, and Mistral-7B-Instruct-v0.3 remain appropriate biomedical/general-purpose comparators for future independent validation [26-28]. The ChatGPT interface also did not expose decoding parameters for exact replication. Finally, exact verification of a claim does not establish biological truth, clinical benefit, or regulatory fitness. It establishes only that the claim is faithful to the evidence and model output declared by the system.

These limitations define the next experiments. The ontology, reference thresholds, evidence schema, attribution rule, and verifier should be frozen and applied to a third prospectively specified cohort. The language interface should then be evaluated on independently authored, genuinely free-form explanations and across multiple general and biomedical LLMs, including direct comparisons with BioMistral, Meditron, and Mistral-family instruction models [26-29]. A direct comparison of unrestricted prose, evidence-cited prose, and constrained dual-output generation (human-readable explanation plus atomic claim object) would test whether schema-constrained generation reduces extraction error [30]. Endpoints should include field-level extraction fidelity, Evidence-ID linking accuracy, unsupported-causality detection, molecular-overstatement detection, and end-to-end verifier sensitivity after LLM structuring. Such evaluation would also align the framework more closely with current medical-imaging AI reporting guidance and lifecycle-oriented trustworthy-AI principles [20-22]. The same architecture can extend beyond glioblastoma to any biomedical AI setting in which model outputs can be linked to addressable patient-specific evidence.

## Conclusion

We introduce evidence-bound reasoning as a verification-centered extension of biomedical explainable AI (XAI). Rather than treating a readable rationale, feature attribution, or model confidence as sufficient evidence of faithful explanation, the framework formalizes patient-specific radiomic measurements as addressable evidence objects and requires explanatory statements to resolve to those objects through a fixed atomic-claim schema. This design turns the relationship between a statement and its declared evidence into something that can be tested directly rather than inferred from narrative plausibility.

The original contributions of the present work are fourfold. First, compared with our earlier neuro-semantic representation [12], we add explicit Evidence IDs, reference-threshold provenance, model linkage, and a deterministic verifier, thereby converting interpretable phenotype tokens into auditable evidence objects. Second, we experimentally separate predictive performance from verifiability: exact claim verification remained 1.000 while controlled ROC AUC was degraded from 0.899 to 0.500, and the same evidence architecture remained available in the weak external MGMT transfer setting. Third, we demonstrate a modular LLM-to-schema interface in which a language model may structure human-readable explanations while final adjudication remains outside the language model. Fourth, we treat cross-cohort transportability as measurable metadata attached to evidence rather than as an assumed property of a semantic label, preserving the distinction between a correctly cited measurement and a universally stable biological representation.

These contributions distinguish the framework from conventional XAI approaches that primarily ask why a model produced an output. Evidence-bound reasoning instead asks whether each claim about that output is traceable to the correct patient, modality, region, descriptor, semantic state, probability, and interpretive scope. The framework is also deliberately distinct from causal AI: feature contributions and radiomic associations are not interpreted as causal effects, and unsupported causal language is rejected as an evidentiary-scope violation. In this sense, the method provides a practical bridge between explainability and auditability without claiming causal inference.

The study does not establish an externally validated MGMT diagnostic model, biological truth, or clinical utility. Its central methodological result is narrower but, we believe, more general: predictive validity, semantic transportability, language-to-schema fidelity, causal scope, and claim verifiability can be represented and evaluated separately. Future work should test this architecture in a frozen third cohort and with independently authored free-form explanations across multiple biomedical and general-purpose LLMs. If those evaluations are successful, evidence-bound reasoning could provide a reusable verification layer for biomedical AI systems in which human-readable explanations must remain explicitly linked to inspectable patient-specific evidence.